\pdfoutput=1

\documentclass[11pt]{article}

\usepackage{acl}

\usepackage{times}
\usepackage{latexsym}

\usepackage[T1]{fontenc}

\usepackage[utf8]{inputenc}

\usepackage{microtype}

\usepackage{enumitem}
\setitemize{noitemsep,topsep=0pt,parsep=0pt,partopsep=0pt}
\definecolor{cite}{rgb}{0.6,0.6,1.0}

\definecolor{todo}{rgb}{1,0.5,0}

\usepackage{booktabs}
\usepackage{graphicx}
\usepackage{float}
\usepackage{tabularx}
\usepackage{soul}

\title{An Inclusive Notion of Text}

\author{Ilia Kuznetsov \and Iryna Gurevych\\ 
Ubiquitous Knowledge Processing Lab (UKP-TUDA) \\
Department of Computer Science and Hessian Center for AI (hessian.AI) \\
Technical University of Darmstadt\\
\texttt{ukp.informatik.tu-darmstadt.de}}

\begin{document}

\maketitle
\begin{abstract}

Natural language processing (NLP) researchers develop models of grammar, meaning and communication based on written text. Due to task and data differences, what is considered text can vary substantially across studies. A conceptual framework for systematically capturing these differences is lacking. We argue that clarity on the notion of text is crucial for reproducible and generalizable NLP. Towards that goal, we propose common terminology to discuss the production and transformation of textual data, and introduce a two-tier taxonomy of linguistic and non-linguistic elements that are available in textual sources and can be used in NLP modeling. We apply this taxonomy to survey existing work that extends the notion of text beyond the conservative language-centered view. We outline key desiderata and challenges of the emerging inclusive approach to text in NLP, and suggest community-level reporting as a crucial next step to consolidate the discussion.

\end{abstract}

\section{Introduction}
\label{sec:intro}

Text is the core object of analysis in NLP. Annotated textual corpora exemplify NLP tasks and serve for training and evaluation of task-specific models, and massive unlabeled collections of texts enable general language model pre-training. To a large extent, natural language processing today is synonymous to text processing.

\begin{figure}[h]
    \centering
    \includegraphics[width=0.8\linewidth]{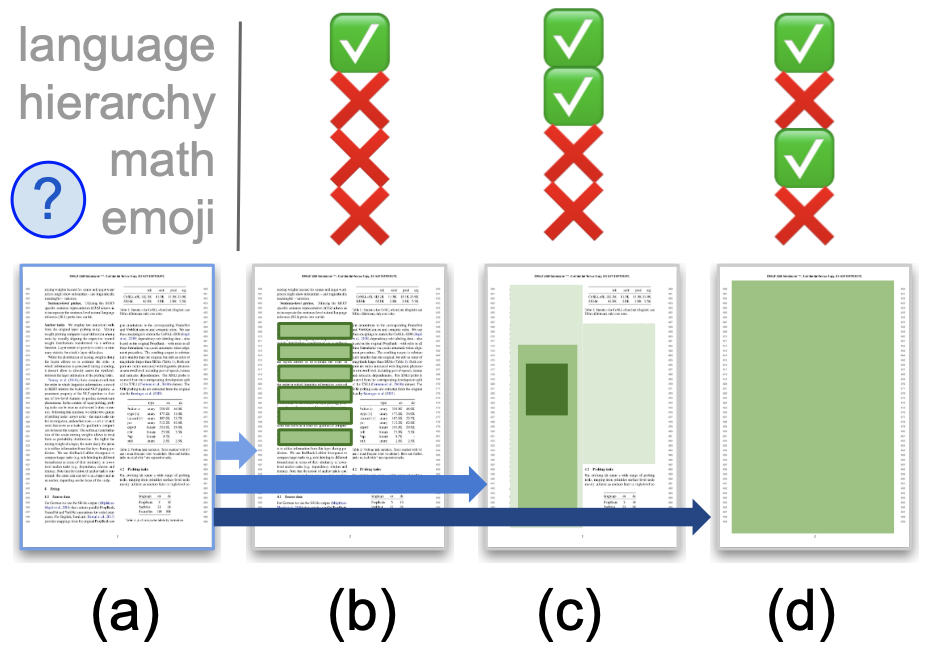}
    \caption{The same textual document (a) can be seen in many ways (b-d) depending on the assumed notion of text: while a syntax researcher might focus on written language (b), a summarization system can use document structure (c), and multimodal applications might use non-linguistic elements like tables (d). Systematically capturing the differences between the assumed notions of text (top) requires a taxonomy of inclusive approaches to text. Such taxonomy is currently lacking.}
    \label{fig:views}
\end{figure}

But what belongs to text? More broadly, what information should be captured in NLP corpora and be available to the models during training and inference? Despite its central role, the notion of text in NLP is vague: while earlier work mostly focused on grammatical phenomena and implicitly limited text to written language, the applied NLP of the past years increasingly takes an \emph{inclusive} approach to text by introducing non-linguistic elements into the analysis. Extensions vary from incorporating emojis to exploiting document structure and cross-document relationships, and apply to all major components of the modern NLP infrastructure, including unlabeled text collections \cite{s2orc}, language models \cite{htlm} and annotated corpora \cite{f1000rd}. The assumption that text in NLP solely refers to written language no longer holds. Yet, as Figure \ref{fig:views} illustrates, a systematic approach to capturing the differences between the assumed notions of text is lacking.

This is problematic for several reasons. From the \textbf{reproducibility} perspective, machine learning assumes similarity between the source and the target distribution -- yet lack of consensus on the notion of text might result in undocumented change of the input representation and degraded performance, even if other common variables like domain, language and task remain unchanged. From the \textbf{modeling} perspective, the notion of text has major influence on task and model design, as it both determines the tasks NLP aims to tackle, and implies what information should be used to perform those tasks. The final argument for studying the notion of text in NLP is \textbf{conceptual}: the capabilities of strong pre-trained Transformer models \cite{bertology} and general-purpose NLP frameworks \cite{allennlp, flair, hftransformers} have led to an explosive growth in NLP beyond traditional, core tasks. The exposure to rich source document types like scientific articles \cite{s2orc} and slides \cite{psed} and the growing influence of multimodal processing \cite{multimod} motivate the use of additional signals beyond written language in NLP. This leads to a general question on the scope of the field: if written language is no longer the sole object of study, what is, and how can it be formally delineated?

Any empirical discipline relies on \emph{operationalization}, which casts observed phenomena into abstractions, allowing us to formulate claims and perform measurements to evaluate these claims. For example, operationalizing sentiment (phenomenon) as a binary variable (abstraction) allows us to a build a claim (\textit{"this review is positive"}) to be evaluated against the ground truth (review rating), and dictates the downstream NLP task design (binary classification). While widely used, this operationalization is limited: alternative notions of sentiment allow making more nuanced claims, fine-grained measurements and precise models.

The same logic applies to text, which affords a wide range of operationalizations, from a character stream \cite{flair} to a rich multimodal graph \cite{f1000rd}. Yet, the typology for describing text use in NLP is lacking. While concurrent proposals address other key properties of NLP models and corpora \cite{metashare, datasheets, datastatements, modelcards} like domain, language, demographics, modality and licensing -- we lack common terminology and reporting schemata for documenting and formally discussing the assumed notion of text. The growth of the field and the high cost of the retrospective documentation underline the urgent need for a lightweight, semi-structured reporting mechanism to account for text use. To address this need, we contribute the following:

\begin{itemize}
    \item A common terminology of text use in NLP (Section \ref{sec:concept});
    \item A taxonomy of text extensions beyond the language-focused approach to text \mbox{(Section \ref{sec:notion})}, based on commonly used sources of NLP data and the current state of the art;
    \item Discussion of the challenges brought by the inclusive approach to text (Section \ref{sec:disc});
    \item A new lightweight semi-structured schema for reporting text use in NLP (Section \ref{sec:quest}).
\end{itemize}

The notion of text is central to NLP, and we expect our discussion to be broadly relevant, with particular merit for the documentation policy, NLP applications, and basic NLP research. The semi-structured reporting as proposed here is a crucial step towards developing formalized documentation schemata \cite{metashare} for describing text use and general formats \cite{nif} to encode non-linguistic information in texts. We encourage the community to adopt our reporting schema, and to contribute to the discussion by suggesting new phenomena to be covered by the taxonomy of inclusive approaches to text.

\section{Terminology}
\label{sec:concept}

\begin{figure}
    \centering
    \includegraphics[width=\linewidth]{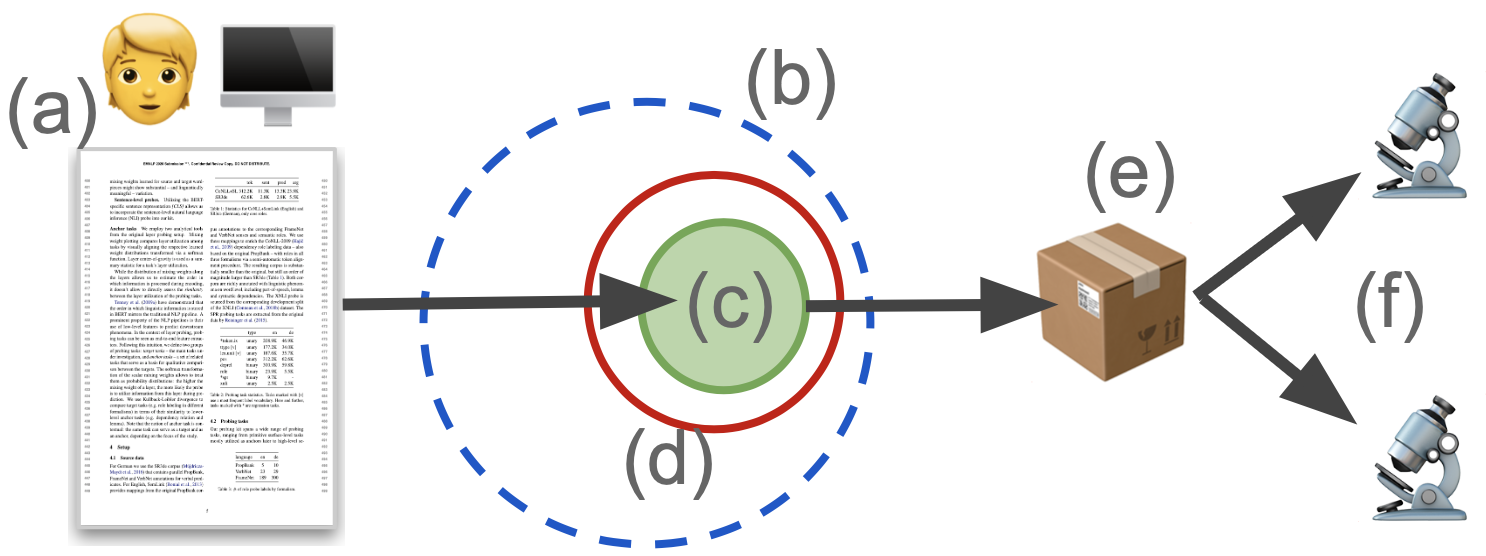}
    \caption{A text is produced in an environment (a) and becomes part of the document space (b) that is sampled (c), often based on source (d). The sample is transformed into NLP artifacts (e) that are potentially reused and further refined across multiple studies (f) to produce further artifacts, etc. This process determines the notion of text assumed by the downstream NLP research and the capabilities of the resulting artifacts.}
    \label{fig:datajourney}
\end{figure}

Textual data available to NLP is a result of multiple processes that determine the composition and properties of texts. To support our discussion, we outline the data journey a typical text undergoes, and introduce common terminology. Figure \ref{fig:datajourney} illustrates our proposed model, and the remainder of this Section provides more details.

\paragraph{Text production.} Every text has been produced by a human or an algorithm with a certain communicative purpose.
Raw text is rarely exchanged; to avoid ambiguity, we use the term \emph{document} for a unit of information exchange\footnote{There are many other kinds of documents, e.g. images, audio or code; here we focus on "textual" documents.}. Documents consist of text along with additional structural and multimodal elements, serialized in a certain \emph{format} and accompanied by metadata. In our definition, textual documents cover a broad spectrum ranging from blog posts, Wikipedia articles and Tweets to dialogue turns and search queries. A few widely used formats are plain text, Markdown, PDF.

\paragraph{Document space.} All textual documents ever produced make up the abstract \emph{document space}. Document space incorporates both persistent textual documents that are stored (e.g. Wikipedia articles), and transient textual documents that only exist temporarily (e.g. search queries). Despite the apparent abundance of textual documents on the Web, a large (if not major) part of the document space is \emph{not} openly available, or is protected from research use by the copyright, privacy and technical constraints.

\paragraph{Sampling and sources} Since capturing the entire document space is not feasible, a \emph{sample} from the subspace of interest is used. Document space can be segmented in a variety of ways, including language, domain or variety \cite{plank16}, creation time, etc. One common way to sample textual documents is based on \emph{source}: documents from the same source often share key characteristics like language variety, text production environment, format and licensing. Some widely used data sources in NLP are Wikipedia, arXiv etc. \cite{wikiated, peerread}.

\paragraph{NLP Artifacts} Sampled textual documents are used to create artifacts, including \emph{reference collections} like BooksCorpus \cite{bookscorpus} and C4 \cite{c4}, and widely reused general-purpose \emph{language models} like BERT \cite{bert} and GPT-3 \cite{gpt3}. 
The notion of text assumed by NLP artifacts is shaped both by the data journey and by the preprocessing decisions during artifact construction. These, in turn, determine how text is operationalized downstream. Due to the differences in how text is produced, sampled and captured, two NLP artifacts might assume very different notions of text. Yet, a framework to systematically capture this difference is lacking.

\section{Prior efforts}

Our proposal draws inspiration from recent efforts in documenting other common properties of machine learning and NLP artifacts. Model cards \cite{modelcards} capture core information about machine learning models including technical characteristics, intended and out-of-scope use and preprocessing details. Data sheets \cite{datasheets} focus on dataset composition, details of the data collection process, preprocessing, distribution and maintenance. In NLP, data statements \cite{datastatements} focus on bias mitigation, detailing key aspects of NLP artifact production such as curation strategy, language variety, demographics of speakers and annotators, speech situation, topic and genre. \citet{dave} propose a formalised checklist documenting risks related to copyright, bias, privacy and confidentiality. Formal proposals are mirrored by community efforts on data repositories like \textit{huggingface datasets} \cite{hfdatasets}; editorial guidelines\footnote{\url{https://aclrollingreview.org/responsibleNLPresearch/}} encourage the authors to report key parameters of NLP artifacts. Related metadata collection initiatives propose schemata for capturing core information about language resources like language, type, license and provenance \cite{metashare}.

While existing approaches to NLP artifact documentation cover a lot of ground, the requirements for documenting the assumed notion of text remain under-specified. Our work is thus complementary to the prior efforts. Our reporting schema (Section \ref{sec:quest}) can be seen as specification of the \texttt{Speech Situation} and \texttt{Text Characteristics} sections of the data statements \cite{datastatements}, and our taxonomy incorporates some previously proposed documentation dimensions like text creation environment \cite{metashare} and granularity \cite{nif}. Unlike most prior approaches, we deem it desirable to document the assumed notion of text at each step of the NLP data journey, including text production tools, document space samples, as well as NLP models and datasets, with a special focus on widely reused reference corpora and pre-trained language models.

\section{Taxonomy of text extensions}
\label{sec:notion}

\subsection{Preliminaries}

We derive our proposal in a bottom-up fashion based on two categories of sources. The text production stage is critical as it determines what information is potentially available to downstream processing; to approximate what information \emph{could be used} by NLP artifacts, we (1) conduct an analysis of four representative document sources widely employed in NLP. On the other side of the data journey are the NLP artifacts, the end-product of NLP preprocessing, modeling and annotation. To approximate what information \emph{is being used} by NLP, we outline the de-facto, conservative approach to text and (2) survey recent efforts that deviate from it towards a more inclusive notion of text.

\paragraph{Sources.} Wikipedia\footnote{\url{https://wikipedia.org}} (\texttt{Wiki}) is a collaborative encyclopedia widely used as a data source for task-specific and general-purpose NLP modeling. BBC News\footnote{\url{https://www.bbc.com/news}} (\texttt{BBC}) represents newswire, one of the "canonical" domains characterized by carefully edited written discourse. StackOverflow\footnote{\url{https://stackoverflow.com}} (\texttt{Stack}) is a question-answering platform that represents user-generated technical discourse on social media. Finally, ACL Anthology\footnote{\url{https://aclanthology.org}} (\texttt{ACL}) is a repository of research papers from the ACL community and represents scientific discourse -- a widely studied application domain \cite{aclant, nlpschol, multicite}. For our analysis we sampled five documents from each of the data sources (Appendix \ref{sec:appsource}): for \texttt{Wiki}, we selected featured articles from five distinct portals to ensure variety; from \texttt{BBC} we selected top five articles of the day\footnote{All documents retrieved on October 4th, 2022}; for \texttt{Stack} we used five top-rated question-answer threads; for \texttt{ACL}, we picked five papers from the proceedings of ACL-2022 available online. Each document was exported as PDF to accurately reproduce the source, and manually annotated for non-linguistic phenomena by the paper authors, with the annotation refined over multiple iterations.

\paragraph{Baseline: Written language.}

The conservative, de-facto approach to text in NLP is "text as written language": parts of source documents that contribute to grammatical sentences are the primary modeling target, whereas non-grammatical elements are considered noise and potentially discarded. This tradition is persistent throughout the history of NLP, from classic NLP corpora \cite{ptb, ontonotes} and core NLP research, to modern large-scale unlabeled corpora used for model pre-training \cite{bookscorpus, wikitext, c4}, language models \cite{bert, gpt3} and benchmarks \cite{glue}. While focus on text as written language is justified for grammatical and formal semantic analysis, for other use cases it proves limiting. In the following Section we survey the emerging inclusive approaches to text that exploit non-linguistic signals to boost the performance and to enable new applications of NLP.

\subsection{Taxonomy overview}

\begin{figure*}
    \centering
    \includegraphics[width=0.87\linewidth]{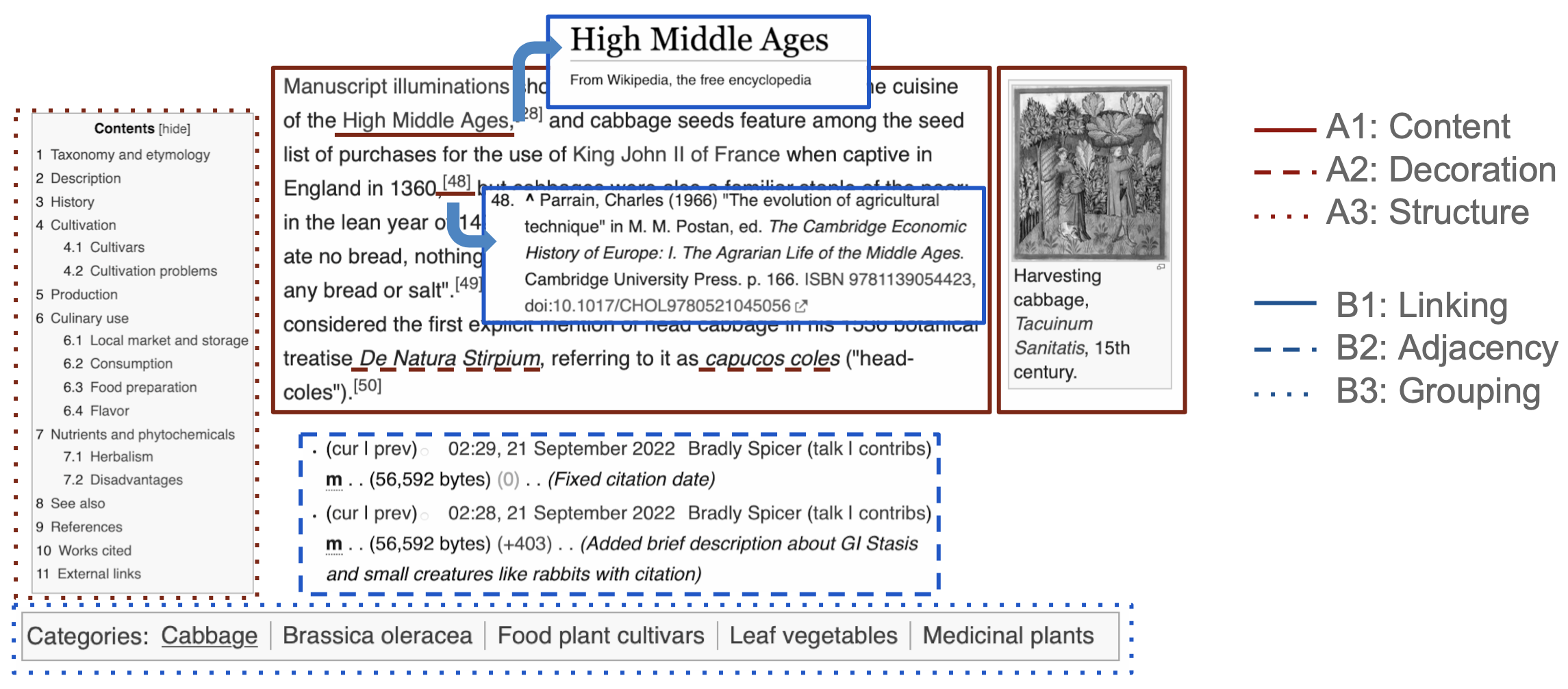}
    \caption{Taxonomy classes in a \texttt{WIKI} document. Besides language, content (A1) includes link anchors and images, decoration (A2) marks code-switching to Latin, the article is structured (A3), linked to external sources via hyperlinks and citations (B1), accompanied by an edit history (B2) and contextualized by its Categories (B3).}
    \label{fig:taxex}
\end{figure*}

Table \ref{tab:tax} summarizes our proposed two-tier taxonomy for describing the inclusive approaches to text. It demonstrates the wide variety of signals available and potentially relevant to NLP processing beyond the conservative, language-centric view. The following sections discuss the taxonomy classes in greater detail, and Figure \ref{fig:taxex} gives examples.

\begin{table*}[]
    \centering
    \begin{tabular}{l|l|l}
        & & examples \\
        \hline
         & A1: Content & written language,
         anchors, 
         math, 
         code, 
         emoji, 
         multimedia\\
        A: Body & A2: Decoration & formatting, color \\
        & A3: Structure & document hierarchy, blocks, page and line numbers \\
        \hline
        & B1: Linking & implicit links, hyperlinks, citations, footnotes \\
        B: Context & B2: Adjacency & comments under post, product and review \\
        & B3: Grouping & tags, document collections, groups \\
        
    \end{tabular}
    \caption{Taxonomy of the inclusive notion of text.}
    \label{tab:tax}
\end{table*}

\subsection{Body}

The first high-level class of our taxonomy encompasses the phenomena related to the main, content-bearing parts of the textual document.

\paragraph{A1: Content.} Our source analysis reveals that naturally occurring textual documents systematically make use of signal systems beyond written language. The examples of non-linguistic information in textual documents include, but are not limited to, emojis, math, code, hyperlink-, citation- and footnote anchors, tables and multimedia, as well as arbitrary numerical and categorical information like scores and ratings (e.g. on \texttt{STACK}). The stance towards such non-linguistic elements of text ultimately determines whether an NLP artifact can represent them in a satisfactory manner, and recent NLP works successfully use non-linguistic elements to their advantage. Applications in sentiment analysis make use of emoji \cite{sentemoji}; recent research addresses text generation based on tables \cite{tabtotext}; \citet{scaffold} use citation anchors for citation intent prediction; \citet{vila}, \citet{markuplm} and \citet{htlm} integrate layout information into language model pre-training, resulting in improved performance across a wide range of tasks. The ability to handle non-linguistic signals is key for NLP applications and motivates careful documentation of text content.

\paragraph{A2: Decoration.} Content is complemented by decoration across all of our sources. Decoration can take the form of font, style, coloring etc. and carries important secondary information, including emphasis, quotation, and signaling \mbox{Structure (A3)}. An important function of text decoration is to mark code-switching between different signal systems, from language change to mathematical notation and code, e.g. on \texttt{STACK} and \texttt{ACL}. Over the past years, decoration received some attention in NLP: \citet{shirani, shiranist} explore the task of emphasis modeling in visual media, \citet{psed} extend it to presentation slides. While humans widely use text decoration, the semantics of decoration are source- and author-dependent and require further systematic investigation.

\paragraph{A3: Structure.} Most naturally occurring textual documents are not a flat, linear text as assumed by commonly used reference corpora, from Penn TreeBank \cite{ptb} to BooksCorpus \cite{bookscorpus}. Instead, the relationships between individual units of content are encoded in document structure. The simplest form of structure is paragraph; longer documents can exhibit a hierarchy of sections; visual proximity is used to include additional content blocks like quotations, definitions, footnotes, or multimedia. In print, textual documents can be organized into pages, columns, lines etc. Explicit document structure is increasingly used in NLP: \citet{scaffold} use sections to help citation intent prediction; \citet{histruct} exploit document structure to aid summarization; \citet{chapterbr} use structure to study the capabilities of long-range language models; \citet{f1000rd} propose Intertextual Graph as a general structure-aware data model for textual documents and use it to support annotation studies and explore how humans use document structure when talking about texts. Document structure is implicitly used in HTML-based pre-training of language models \cite{htlm}, yielding superior performance on a range of tasks, and enabling new pre-training strategies; a separate line of study is dedicated to the analysis of visual document layouts \cite{vila}. The lack of a common approach to formalizing document structure calls for systematic reporting of what structural elements are available in sources, and how document structure is represented and used in NLP.

\subsection{Context}

The second high-level class of our taxonomy pertains to context. Every text is written and read in the context of other texts, and the ability to capture and use context is a key property of NLP artifacts.

\paragraph{B1: Linking.} The first major contextualization mechanism is explicit linking -- a marked relationship between an anchor text and a target text \cite{f1000rd}. Linking is crucial to many text genres and is found throughout the document sources considered in our analysis. An intra-document link connects two elements within one textual document (e.g. reference to a chapter or footnote), while a cross-document link connects elements in different documents (e.g. hyperlinks and citations). Links differ in granularity of their anchors and targets: the same \texttt{Wiki} page can cite its sources on the level of individual sentences (sentence to document) and as a list for further reading (document to document); a research article from \texttt{ACL} can reference a particular statement in a cited work (sentence to sentence). A few recent works tap into the narrow context for both task-specific and general-purpose modeling: \citet{ape} investigate the relationships between peer reviewer comments and author rebuttals; \citet{specter} use information from the citation graph to create better neural representations of scientific documents; \citet{bugert} exploit hyperlinks to generate cross-document event coreference data; \citet{cdlm} show that jointly encoding a document and its near context improves performance on tasks like cross-document coreference resolution and multi-hop question answering; \citet{f1000rd} and \citet{disapere} jointly model cross-document relations between manuscripts, peer reviews, revisions and author responses. The availability and use of cross-document links are key properties of textual documents and NLP artifacts to be documented.

\paragraph{B2: Adjacency.} In addition, textual documents can be related by adjacency; common examples include commentaries attached to the main text, discussion thread replies, copyright notices and prefaces, or peer reviews and the submissions they discuss. Contextualization by adjacency is at play in the NLP study of discussion threads \cite{jamison}, peer reviews \cite{doesmyrebuttal, ape, disapere}, etc. Temporal adjacency is a special case where a textual document exists in the context of its previous and future revisions, and is a key feature of document sources like \texttt{Wiki}; edit histories have been widely used in NLP as a modeling and annotation target \cite{wikiedit, f1000rd, fruit, peer, newsedits}. Like linking, adjacency is a rich, naturally occurring type of contextualization.

\paragraph{B3: Grouping} Finally, a textual document can be contextualized by the region of the document space that it belongs to: a \texttt{Wiki} page exists in the context of other pages belonging to the same portal; a \texttt{BBC} article is positioned along the other articles of the same day or topic. Group context both provides the expected common background for text interpretation and sets the standards for the composition of individual documents. Group context plays key role in designing discourse segmentation schemata \cite{argzon, ampere, f1000rd, disapere}, can yield natural labels for text classification, and has been used to augment language models \cite{cdlm}.

\subsection{Remarks}
\label{sec:remarks}
\paragraph{Completeness.} Our taxonomy serves as the first attempt at capturing the notion of text used in NLP in a structured manner. While we believe that the high-level taxonomy given here is comprehensive, due to our focus on textual documents we do not incorporate further divisions related to multimedia content (e.g. we do not distinguish between images and graphics, although such distinction could be of interest for some applications). As more sources and NLP artifacts are documented, new lower-level taxonomy classes are likely to emerge.

\paragraph{Interactions.} The proposed taxonomy dimensions are not orthogonal and do interact: for example, group context (B3) can influence document structure (A2) and decoration standards (A3); in turn, decoration is widely used to signal document structure and linking (B1); the presence of adjacent context (B2) can affect the level of detail in the content (A1). The existence of such inter-dependencies motivates joint documentation and analysis of the different aspects of text even if a conservative notion of text is adopted in the end.

\section{Additional considerations}
\label{sec:disc}

\subsection{Interoperability and generalization} A great advantage of the conservative, written-language-only view on text is wide interoperability and generalization: any textual document -- from scientific articles to Tweets -- can be reduced to written language. This makes it possible to apply a BERT model trained on books to a question-answering prompt and expect non-trivial performance, and enables reuse of text processing frameworks and annotation tools. Yet, such reduction leads to substantial information loss and bears the danger of confounding due to the interactions between different aspects of text and the text body. While isolated efforts towards inclusive notion of text exist, we are not aware of general approaches that would allow capturing different aspects of text in a systematic manner across domains and document formats. While arriving at a universal, general inclusive notion of text for NLP might not be feasible, we believe that reflecting on the generalization potential of non-linguistic textual elements is the first step in this direction.

\subsection{Impact of production environment}

Text production environment plays a key role in what information can be captured by the textual document, which, in turn, determines the capabilities of the downstream NLP artifacts. While a sophisticated text editing interface promotes the use of decoration, non-linguistic content, structure and linking, a plain text input field does not. Moreover, the regulating documents and norms that accompany text production have a profound effect on text composition: for example, in addition to common expectations of a scientific publication, \texttt{ACL} provides document templates, sets page limits and often enforces obligatory structural elements e.g. reproducibility and limitation sections; \texttt{Wiki} is supplied with extensive general and portal-specific guidelines, as well as strict formatting requirements enforced by the community; similar mechanisms are characteristic of most other sources of textual data. Finally, the environment might determine the availability of adjacent and group context during text production. Despite its crucial role, we are not aware of NLP studies that investigate the effect of the production environment on the resulting texts, and believe that our taxonomy can serve as a viable scaffolding for such studies.

\subsection{Implications}

\paragraph{Efficiency.} Computational demands of NLP research are a growing concern \cite{strubell}. It remains unclear how the transition to inclusive treatment of textual documents might affect the efficiency of NLP models. Modeling additional aspects of text might require more parameters and increase the computational demands; yet, the synergies between different aspects of text might allow NLP models to converge faster during training. We are not aware of NLP studies that systematically investigate the effects of inclusive approach to text on training of NLP models, and believe that this question requires further scrutiny.

\paragraph{Ethics.} Recent years are marked by increased attention to the ethics of NLP research, broadly including the issues of privacy, confidentiality, licensing and bias \cite{datastatements, dave, 3yw}. While some types of information beyond written language do not constitute a threat as they are openly accessible in the source textual documents (e.g. textual content A1, decoration A2 and structure A3), others are potentially harmful: precise details of text production might impact privacy, and inclusion of certain contexts (e.g. edit histories, B2) might expose NLP artifacts to false and incomplete information. We are not aware of systematic NLP research into what types of non-linguistic information about textual documents are safe to store and report.

\paragraph{Methodology} Current NLP methodology is tailored to a conservative approach to text -- from commonly reported dataset statistics (e.g. number of words) to modeling objectives and evaluation metrics. The transition towards an inclusive notion of text calls for a careful revision of the NLP practice. Dataset statistics might include information like the number of figures and tables (A1) or structural information on intra-document (A3) and inter-document (B1-3) level. Pre-trained language models would need to process new types of content, structure and context. Evaluation metrics would need to take into account the new signals. In addition, machine learning models are prone to heuristic behavior \cite{artifacts} -- and besides providing a useful training signal, inclusive notion of text might introduce spurious cues that the models would exploit. Future research must determine the optimal ways to operationalize the inclusive approaches to text in NLP.

\section{Reporting}
\label{sec:quest}

An inclusive approach to text is an emerging trend in NLP that demands systematic study. While preparing this work, it became evident that the lack of systematic reporting limits the meta-analysis of text use in NLP. In line with related documentation efforts, here we propose a simple, semi-structured mechanism for reporting text use. In the short term, such reporting would make it easier to gauge the capabilities of data sources and NLP artifacts, increase community awareness on what aspects of text are represented and used, and allow aggregation of text use information from different studies. In the long term, it would help the community develop standards for applying the inclusive approach to text and formally documenting text use, and allow informed development of general data models and formats \cite{nif} to facilitate interoperability between NLP artifacts that adopt an inclusive approach to text.

\subsection{\underline{Schema}}

As our proposed taxonomy is subject to extension, and to keep the reporting effort low, we formulate the proposed reporting schema as a set of open-ended questions guided by examples in Table \ref{tab:tax}, in the spirit of short-form data statements by \citet{datastatements}. We encourage the reporters to complement it with new categories and phenomena if necessary. For each NLP study that uses or creates textual documents or NLP artifacts, we propose to include the following information into the accompanying publication:

\begin{itemize}
    \item \textbf{Body}: Does the source, format, dataset, model or tool incorporate or use any information apart from written language, including non-linguistic content, decoration and structure? 
    \item \textbf{Context}: Does the source, format, dataset, model or tool incorporate or make use of additional context beyond single document, including by linking, adjacency or via group context? If yes, what is it and how is it used?
\end{itemize}

In addition, for text document sources and interactive NLP models we propose to document the \textbf{production environment}: How are the documents produced, including guidelines, software and hardware used? Are the documents single-authored or written collaboratively? How can these factors influence text body and context? Optionally, we invite researchers to reflect upon the implications of their approach to text for \textbf{generality}, \textbf{efficiency}, \textbf{ethics} and \textbf{methodology}. Is the newly introduced signal widely used across textual documents? Does it add computational overhead or help reduce computational cost? Can new information lead to bias, privacy risks or promote heuristic behavior? Does the selected methodology take the non-linguistic nature of the new information into account? 

\subsection{Example and Implementation} 

To illustrate the intended use of the proposed schema, Appendix \ref{sec:appexample} provides example documentation for a textual source (StackOverflow). We note that despite the brevity, short form and potential incompleteness, this kind of documentation is highly informative as it both allows to quickly grasp the notion of text assumed by a data source or artifact, and to aggregate this semi-structured information across different kinds of NLP studies in the future.

Unlike prior efforts that focus on documenting datasets and models separately, our schema applies to all stages of the NLP data journey, from data sources to NLP artifacts, including reference corpora, labeled corpora, preprocessing tools, pre-trained and end-task models and applications. The schema can be incorporated into the data statements and editorial guidelines and used to extend prior metadata documentation proposals \cite{metashare} and data repository submission forms \cite{hfdatasets}. 

We encourage the community to make use of this low-effort mechanism as a step towards better interoperability of NLP artifacts and the systematic study of the inclusive notion of text. We specifically highlight the need for documenting commonly used \underline{sources of textual documents}; this will provide the NLP community with a better picture of the document space. We deem it equally important to document \underline{pre-trained language models} and \underline{reference corpora}, since their capabilities have a major effect on downstream NLP modeling and applications. This would allow us to gauge how far NLP is from accurately modeling the document space, and will highlight the gaps future work would need to address on the way towards a generally applicable inclusive approach to text.

\section{Conclusion}

Text plays the central role in NLP as a discipline. But what belongs to text? The rise in applications of NLP to non-linguistic tasks motivates an inclusive approach to text beyond written language. Yet, the progress so far has been limited to isolated research efforts. As NLP ventures into new application areas and tackles new tasks, we deem it crucial to document the notion of text assumed by data sources and NLP artifacts. To this end, we have proposed common terminology and a two-tier taxonomy of inclusive approaches to text, complemented by a widely applicable reporting schema. We hope that our contributions and discussion help the community systematically approach the change of NLP scope towards more accurate modeling of text-based communication and interaction.

\section*{Limitations} Our proposed taxonomy is subject to extension, and we expect new phenomena to be included into its scope as the field progresses and as more document sources are considered. Using a taxonomy as an organizational basis for the proposed schema is dictated by our aim to keep the schema simple. The design of future, formalized reporting schemata might adopt an onthology-based approach as it affords more flexibility, and take into account interoperability with the existing proposals in the linked open data community \cite{nif}.

While source analysis is only one of our contributions and is thus limited in scope, we have observed that increasing the number of documents from the \emph{same} source yields diminishing value: if a source uses a certain non-linguistic textual element, it does so consistently. This suggests that the future qualitative studies of document sources used in NLP should be conducted in a breadth-first fashion, with few documents samples from many sources, unless quantitative measurement is desired (e.g. \textit{"how often do Wikipedia authors use text formatting"}) or unless a source is known to accommodate a wide variety of document types with different publication and formatting standards. 

We do not provide specific details on documenting the text production environment, which represents a promising future research avenue. The study of how the texts in NLP are created is a critical research direction: due to the increased applied use of pre-trained generative language models, documenting the text form and origin is a pressing need.

Our discussion stresses the overall need for more careful handling of terminology in NLP. In this work we chose the term "text" to refer to the object of study in NLP -- hence an approach that incorporates non-linguistic elements into text is considered "inclusive". We note that "text" itself is an overloaded term associated with writing on one hand, and text as a format on the other hand; from a cross-disciplinary perspective, e.g. in semiotics, a musical piece or an advertisement would be termed "text" as well. An alternative terminology would use "document" instead of "text" -- however, we have opted against this choice, as document can be non-textual (e.g. images, spreadsheets), carries certain implications on length, structure and standalone nature (“document-level NLP"), and comes with its own cross-disciplinary connotations. As NLP progresses methodologically and interacts with other disciplines, we deem it plausible that a more precise terminology will emerge.

\section*{Acknowledgements} This study is part of the InterText initiative\footnote{\url{https://intertext.ukp-lab.de}} at the UKP Lab. The study has been funded by the LOEWE Distinguished Chair “Ubiquitous Knowledge Processing” (LOEWE initiative, Hesse, Germany) and co-funded by the European Union (ERC, InterText, 101054961). Views and opinions expressed are however those of the author(s) only and do not necessarily reflect those of the European Union or the European Research Council. Neither the European Union nor the granting authority can be held responsible for them.

\bibliographystyle{acl_natbib}
\bibliography{emnlp2022}
\cleardoublepage
\appendix

\section{Documentation example: text source}
\label{sec:appexample}

\paragraph{StackOverflow} hosts three main types of textual documents: questions, answers and commentaries. \underline{(A) Body}: documents are richly formatted, include multiple content types (text, code, math, images) and decoration (emphasis, code-switching, links). Documents are associated with additional metadata, author and creation/edit time; questions and answers are assigned a rating (number of votes), questions are tagged. Basic structure is present: questions and answers can be logically structured; questions are titled; yet, commentaries are usually short and not structured. \underline{(B) Context}: linking is used throughout, mostly via hyperlinks, both to the documents on the platform and to external documents; questions, answers and commentaries are related by adjacency; revision histories are available for questions and answers; questions are grouped via tags, and answers and commentaries are grouped by question.  \underline{Production environment}: questions and answers are entered via a UI based on Markdown\footnote{\url{https://stackoverflow.com/editing-help}}, that supports formatting, structuring, lists, links, code and block inserts, and table formatting. The question submission form additionally includes a title and a tag field. While posting the answer, the user has direct access to the question, previous answers and commentaries. Guidelines for asking and answering questions are available\footnote{\url{https://stackoverflow.com/help/how-to-ask}} and enforced both by explicit moderation and by the community.

\section{Source documents}
\label{sec:appsource}

\begin{table}
    \centering
    \begin{tabular}{l|cccc}
         & WIKI & BBC & STACK & ACL \\
         \hline
         \textbf{A Body} & & & & \\ 
         \hline
         A1 Content & & & & \\ 
         - math & yes & no & no & yes \\ 
         - code & no & no & yes & yes \\ 
         - hyperlinks & yes & yes & yes & yes \\ 
         - citations & yes & no & no & yes \\ 
         - footnotes & yes & no & no & yes \\
         - images & yes & yes & yes & yes \\
         \hline
         A2 Decoration & & & & \\ 
         - font & yes & no & yes & yes \\ 
         - style & yes & yes & yes & yes \\ 
         \hline
         A3 Structure & & & & \\ 
         - paragraphs & yes & yes & yes & yes \\ 
         - sections & yes & yes & yes & yes \\ 
         - blocks & yes & yes & no & yes \\ 
         - pages & no & no & no & yes \\ 
         - columns & no & no & no & yes \\ 
         \hline
         \textbf{B Context} & & & & \\ 
         \hline
         B1 Linking & yes & yes & yes & yes \\ 
         \hline
         B2 Adjacency & yes & yes & yes & no\\ 
         \hline
         B3 Grouping & yes & yes & yes & yes \\ 
    \end{tabular}
    \caption{Non-linguistic elements of text by data source, "yes" -- encountered in at least one document from the study sample.}
    \label{tab:srcex}
\end{table}

Table \ref{tab:srcex} summarizes our source analysis. Note that it serves an illustrative purpose and should be used neither as a comprehensive list of non-linguistic phenomena (see Section \ref{sec:notion} instead), nor as a comprehensive documentation of the  data sources: if substantially more documents were considered, mathematical notation would be eventually found in \texttt{STACK}, a \texttt{WIKI} article would eventually feature a code snippet, and an eventual \texttt{ACL} paper would be accompanied by an adjacent erratum or a peer review. The list below enumerates the documents used in our study, retrieved on October 4\textsuperscript{th}, 2022. 

\textbf{WIKI}
\begin{itemize}
\item \url{https://en.wikipedia.org/wiki/Euclidean_algorithm}
\item \url{https://en.wikipedia.org/wiki/Cabbage}
\item \url{https://en.wikipedia.org/wiki/1689_Boston_revolt}
\item \url{https://en.wikipedia.org/wiki/Abdication_of_Edward_VIII}
\item \url{https://en.wikipedia.org/wiki/243_Ida}
\end{itemize}

\textbf{STACK}
\begin{itemize}
\item \url{https://stackoverflow.com/questions/477816/which-json-content-type-do-i-use}
\item \url{https://stackoverflow.com/questions/5767325/how-can-i-remove-a-specific-item-from-an-array}
\item \url{https://stackoverflow.com/questions/6591213/how-do-i-rename-a-local-git-branch}
\item \url{https://stackoverflow.com/questions/348170/how-do-i-undo-git-add-before-commit}
\item \url{https://stackoverflow.com/questions/1642028/what-is-the-operator-in-c}
\end{itemize}

\textbf{BBC}
\begin{itemize}
\item \url{https://www.bbc.com/news/business-63126558}
\item \url{https://www.bbc.com/news/world-latin-america-63126159}
\item \url{https://www.bbc.com/news/world-europe-63119180}
\item \url{https://www.bbc.com/news/world-australia-63126430}
\item \url{https://www.bbc.com/news/world-asia-india-63127202}
\end{itemize}

\textbf{ACL}
\begin{itemize}
\item \url{https://aclanthology.org/2022.acl-long.6.pdf}
\item \url{https://aclanthology.org/2022.acl-long.7.pdf}
\item \url{https://aclanthology.org/2022.acl-long.8.pdf}
\item \url{https://aclanthology.org/2022.acl-long.9.pdf}
\item \url{https://aclanthology.org/2022.acl-long.10.pdf}
\end{itemize}


\end{document}